# SortingEnv: An Extendable RL-Environment for an Industrial Sorting Process

Tom Maus, [a] , Nico Zengeler, [b] and Tobias Glasmachers [c]

*Ruhr-University Bochum, Universitätsstraße 150, 44801 Bochum, Germany*

[a] Corresponding author: tom.maus@ini.rub.de
[b] nico.zengeler@ini.rub.de
[c] tobias.glasmachers@ini.rub.de

**Abstract.** We present a novel reinforcement learning (RL) environment designed to both optimize industrial sorting systems and study agent behavior in evolving spaces. In simulating material flow within a sorting process our environment follows the idea of a digital twin, with operational parameters like belt speed and occupancy level. To reflect real-world challenges, we integrate common upgrades to industrial setups, like new sensors or advanced machinery. It thus includes two variants: a basic version focusing on discrete belt speed adjustments and an advanced version introducing multiple sorting modes and enhanced material composition observations. We detail the observation spaces, state update mechanisms, and reward functions for both environments. We further evaluate the efficiency of common RL algorithms like Proximal Policy Optimization (PPO), Deep-Q-Networks (DQN), and Advantage Actor Critic (A2C) in comparison to a classical rule-based agent (RBA). This framework not only aids in optimizing industrial processes but also provides a foundation for studying agent behavior and transferability in evolving environments, offering insights into model performance and practical implications for real-world RL applications.

## INTRODUCTION

### Reinforcement Learning and Industry

In the rapidly evolving field of machine learning, reinforcement learning has emerged as a powerful paradigm for optimizing decision-making through trial and error with minimal human intervention [1]. Much of the progress in RL has been driven by benchmarks set in virtual gaming environments [2]. However, gaming benchmarks often lack the applicability to real-world problems in terms of complexity, stochasticity, and safety constraints that are usually present in industrial setups. In real-world industrial settings, the repercussions of poor decisions can be immediate and severe, requiring a more cautious approach to exploration [3–5].

Industrial environments are often dynamic and error-prone, with frequent changes in processes, machinery, and sensors. Traditionally programmed control systems struggle to adapt seamlessly to these changes in real time [6]. Dynamic action and observation spaces also pose significant conceptual challenges for RL in industrial applications, underscoring the need to tailor swiftly evolving RL solutions for the complexities of industrial applications[3].

### Our Framework: SortingEnv

We found a lack of industrial RL environments that are both publicly available and can be extended to challenge agents with increasing complexity. To bridge this gap, we propose a novel RL environment designed for industrial sorting systems. The industrial sorting process is a critical operation in many manufacturing and recycling facilities. The environment used in our framework is inspired by a common sorting setup, as e.g. described by Kroell et al. in a recent publication [7].

It is designed to simulate material flow and sorting dynamics while utilizing operational parameters such as belt speed and occupancy levels to safely determine optimal control tradeoffs, following the idea of a digital twin. It models the flow of material between multiple compartments, such as "Input", "Conveyor Belt", "Sorting Machine" and "Storage" (see Fig. 1). We present two versions of our environment:

The basic version focuses on discrete adjustments of the belt speed. The agent must learn to dynamically adjust belt speed by observing belt occupancy to maintain high sorting accuracy. The advanced version builds on top of the basic environment by introducing multiple sorting modes and more detailed material composition observations.

This environment provides additional dimensions for the agent's decision-making process, allowing for more sophisticated strategies. Modes include basic sorting, positive sorting, and negative sorting, which are described in Section 3.3.

We detail the state update mechanisms, reward functions, and observation spaces for both environments, illustrating their roles in training and benchmarking RL agents. We further use state-of-the-art algorithms such as PPO, DQN, and A2C [2, 8, 9] in comparison to a classical rule-based agent (RBA) to present performance evaluations.

In conclusion, our framework not only aids in optimizing industrial sorting processes but also serves as a foundation for future research on agent behavior and adaptability in evolving environments, providing a basis for work in the fields of applied Transfer Learning, Meta-Learning, and Continual Learning [10].

The structure of this paper is as follows: Section 2 reviews relevant literature and establishes the motivation for our work. Section 3 describes the industrial sorting task that inspired our study and formulates the problem as an RL task, explaining the design choices and implementation details of our environments. Section 4 presents the setup and results of our experiments and benchmarking with baseline methods. Finally, Section 5 concludes with a discussion of our findings.

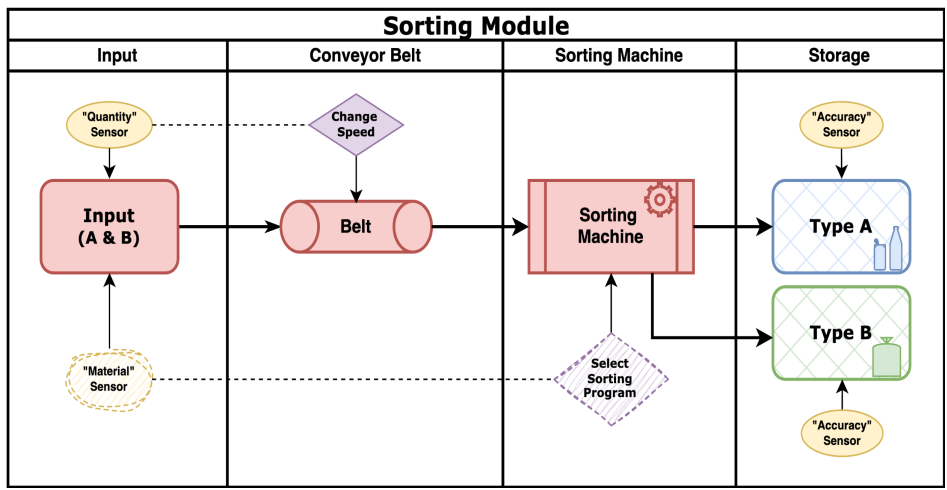

**FIGURE 1.** General overview of the Sorting Environment with the four compartments "Input", "Belt", "Sorting Machine" and "Storage". The simple environment is described by the upper elements for sensors (yellow) and actions (purple). The configuration for the advanced environment is shown by additional sketched elements below, for another sensor (yellow) and action (purple).

## RELATED RESEARCH

This section provides an overview of existing work in the fields of evolving industrial environments, reinforcement learning, and available environments for benchmarking. This review contextualizes our research within the broader academic landscape, highlighting the novelty and relevance of our contributions.

### Handling Evolving Industrial Environments

Industrial facilities often undergo setup changes, such as adding new sensors or updating machine elements to newer, more advanced versions. Traditional methods, such as rule-based models, rely on predefined rules that must be modified to manage upgrades [6]. However, these methods lack the flexibility and adaptability required for dynamic and evolving environments, which may show a considerable amount of sensor noise as well [11]. RL offers a powerful alternative, capable of handling complex, unpredictable environments by learning optimal strategies through interaction with the environment [1]. Unlike Adaptive Control (AC), which is effective in well-understood processes with established models, RL excels in scenarios where no precise model is available, providing opportunities for innovation and adaptation in these industrial contexts [12].

# Reinforcement Learning in Industrial Applications

An RL agent operates by interacting with an environment, taking sequential actions, observing the resulting states, and receiving feedback in the form of rewards, which it tries to maximize [1]. This approach enables the development of agents capable of solving complex tasks without explicit programming for each task. Algorithms such as PPO or DQN can optimize various industrial processes by learning from interactions with the environment and making decisions that maximize a reward function [2, 8].

The application of RL in various industrial sectors, particularly in optimizing control processes thus presents an exciting frontier for exploration. In previous studies, the significant potential of RL in real-world applications has been described, mentioning work on higher efficiency in data centers, gas turbines, and fusion reactors [4], or manufacturing [13] and process optimization [5]. In supply chain management, RL can optimize inventory management, reduce costs, and improve efficiency [14]. Combining these modern methods promises to lower production costs, save energy, and improve product quality. Zhang et al. [15] described the integration of digital twins with RL, demonstrating their potential in various industrial contexts. A recent study presented a digital twin system for autonomous process control using deep reinforcement learning together with supervised learning to both improve model accuracy and optimize operations [16].

Despite some promising applications, the broader adoption of RL in the industry still faces significant hurdles, such as data scarcity, safety constraints, and the dynamic nature of operational environments. So, due to its relatively nascent development and the lack of standardized industrial benchmarks, RL's integration into broader industrial applications remains limited compared to supervised learning [4].

# Benchmark-Environments and Alternatives

As most of the classical RL environments (e.g. [2]) lack the complexity or realism demanded for industrial applications, a critical need for RL benchmarks that can accurately simulate and extend to real-world industrial scenarios arises. To address this, multiple environments have been introduced, bridging the gap between theoretical models and practical applications. Existing frameworks, like the "Industrial Benchmark" [17], often show limited customizability and fail to meet the evolving needs of industrial RL research [4]. The real-world RLsuite by Dulac-Arnold et al. [3] adapts simulated environments to approximate real-world conditions, but it is far from depicting an actual industrial setup. Pendyala et al. [4] recently introduced "ContainerGym", a benchmark designed for a specific industrial application that involves complex resource allocation tasks. This benchmark supports the testing and refinement of advanced RL algorithms like PPO, TRPO, and DQN. However, this task differs significantly from many industrial tasks, as it involves long-term waiting and infrequent actions. Further, its design is not intended to be easily extendable for more complex, conceptual upgrades.

The scarcity of open-source options that can provide realistic, customizable, and extendable testing environments remains a major obstacle to test RL in industry. This gap underscores the pressing need to develop more adaptable and realistic RL benchmarks that can effectively facilitate the translation of RL strategies into successful industrial applications, thereby driving innovation and efficiency across various sectors.

# ENVIRONMENT

In this section, we describe the implementation details of our environment (SortingEnv) tailored for an industrial use case, utilizing the Gymnasium framework in version 0.29.1 [18]. Gymnasium provides a standardized API for creating and working with RL environments, which facilitates experimentation and benchmarking. We present two variants of the environment: a basic version for classical training and an advanced version to provide a baseline for future experiments on evolving setups. The code for these environments can be found in the associated GitHub repository[1].

# The Basic Environment

The basic environment simulates an industrial sorting system for material classification and separation. Our objective is to model the dynamics of material flow through various stages, capturing the impact of operational

---

parameters such as belt speed and material occupancy on sorting accuracy. The system is designed to handle two types of materials, denoted as material A and B, which are sorted using a conveyor belt- and sorting machine setup.

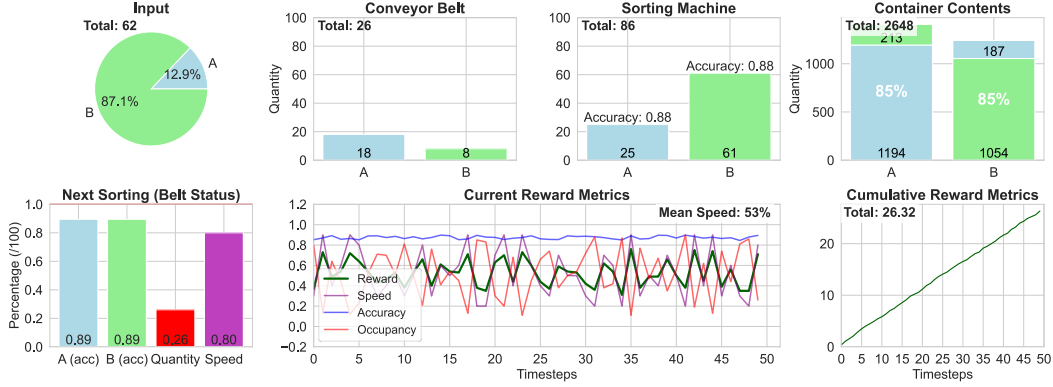

**FIGURE 2.** The dynamic dashboard for the simulations. In this case, it presents data from the basic environment with random input and actions chosen by a DQN agent. The input (upper first) presents the current total amount of material introduced to the system and its distribution. The conveyor belt (upper second) and the sorting machine (upper third) show the current corresponding distribution of material, with the latter also showing the sorting accuracy for the material in the press. The container contents (upper fourth) depict the current amount of correct and falsely classified materials for both types A (blue) and B (green). The current accuracy (acc) per material on the belt, relative quantity of material on the belt, and speed of the belt are shown in the lower left plot. The fluctuations of the belt speed, the accuracy of the material on the belt, the belt occupancy ("quantity"), and the resulting reward per time steps are shown in the lower middle plot. Finally, the cumulative reward of the system is shown on the lower right.

Three main stages characterize material flow in the sorting environment (see Fig. 2):

1. **Input Stage**: The raw material input consists of a randomized mixture of Material A and Material B. This mixture is represented by the vector $I = [I_A, I_B]$, where $I_A$ and $I_B$ denote the quantities of material A and B, respectively. The total amount of input material is:

$$I_{total} = I_A + I_B \tag{1}$$

2. **Belt Stage**: The input material is transferred onto the conveyor belt, described by the vector $B = [B_A, B_B]$, where $B_A$ and $B_B$ represent the quantities of material A and B on the belt. The belt occupancy is calculated as:

$$O_{Belt} = \frac{B_A + B_B}{100} \tag{2}$$

3. **Sorting Stage:** The material on the conveyor belt is then sorted by the sorting machine into separate containers for material A and B. The quantities of sorted materials are denoted by $S = [S_A, S_B]$. The sorting accuracy is defined as $\alpha$, which represent the proportion of correctly sorted material A and B.

The accuracy $\alpha$ of the sorting process for materials A and B is influenced by both belt speed $v$ and occupancy $O_{Belt}$. The maximum occupancy for achieving highest accuracy is determined by predefined occupancy limits $O_{v\_max}$ per belt speed (see Fig. 3, top left). If the belt occupancy $O_{Belt}$ is within this limit for the given speed $v$, the accuracy is set to 1.0. Otherwise, the excess occupancy is calculated by subtracting the occupancy limit from the current belt occupancy. We modeled a linear decrease for the accuracy, inspired by Kroell et al [7], scaled by an abatement rate $\lambda$ (see Fig. 3). To account for variability, a noise factor, uniformly distributed within a predefined range (e.g. 10-15%), is subtracted from the calculated accuracy. The final accuracy for both materials is then clamped between 0.0 and 1.0 to ensure it remains within valid bounds. Thus, the sorting accuracy dynamically adjusts based on operational conditions while incorporating stochastic elements to simulate real-world variability.

$$\alpha = \begin{cases} 1 - \text{Noise} & \text{if } O_{Belt} \leq O_{v\_max} \\ \alpha_{>\text{limit}} = 1 - (O_{excess} * \lambda) - \text{Noise} & \text{if } O_{Belt} > O_{v\_max} \end{cases} \tag{3}$$

A threshold (e.g. 0.7) for the minimum acceptable accuracy can be set to ensure quality standards (see Fig. 4, black-colored areas). The sorting process then updates the quantities of correctly and incorrectly sorted materials based on the current sorting accuracy:

$$S_A = \alpha \times B_A + (1 - \alpha) \times B_B \qquad (4)$$

$$S_B = \alpha \times B_B + (1 - \alpha) \times B_A \qquad (5)$$

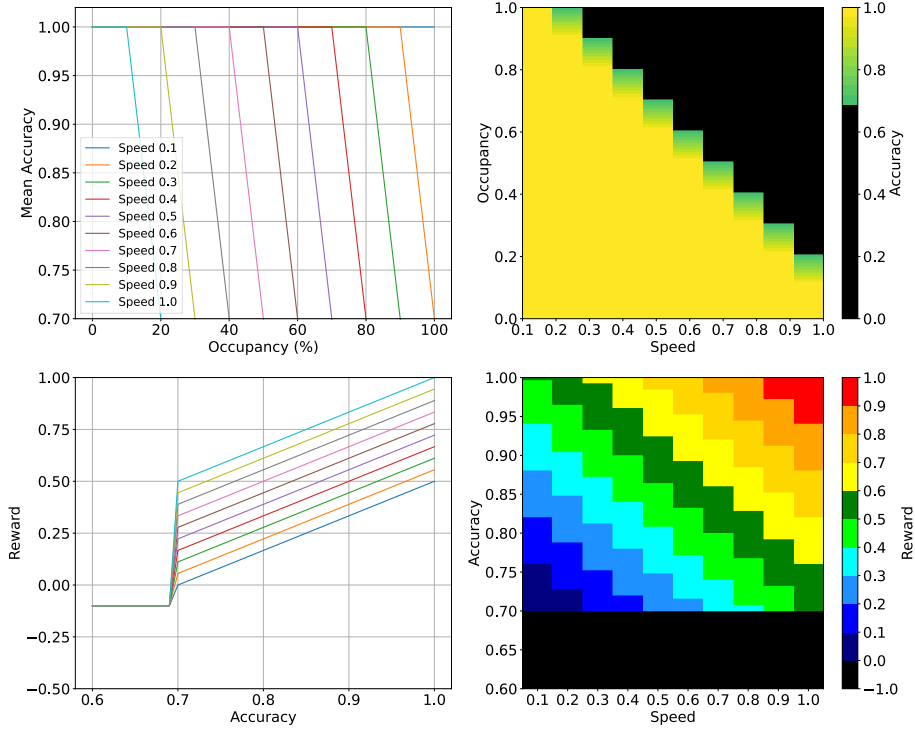

**FIGURE 3.** The plots present the dynamics of the basic sorting environment, focusing on belt speed, occupancy, accuracy, and reward. Accuracies decrease linearly with an abatement rate of $\lambda = 3$. The top left plot shows that mean accuracy decreases linearly after a threshold per speed, responding to increasing occupancy and indicating a trade-off between speed and accuracy. The top right plot depicts accuracy gradients based on speed and occupancy, revealing that lower speeds maintain high accuracy across various occupancy levels. When falling below the threshold of 0.7 (black), a penalty is applied. The bottom left plot illustrates the rewards that can be achieved per combination of speed and accuracy, when above the mentioned threshold. The bottom right plot visualizes a sketched distribution of rewards based on speed and accuracy, showing that the highest rewards are achieved at both high speeds and accuracies.

Incorrectly sorted materials are accounted for as they are relevant in evaluating the systems performance. The operators' goal is to obtain high purity (via "accuracy") in the storage containers while ensuring the most efficient throughput ("speed") of the sorting process. The purity $P$ of the containers is defined as the ratio of correctly sorted material to the total material in the containers ("Precision"), where $S_{A,True}$ and $S_{B,True}$ are the quantities of correctly sorted material A and B, respectively.

$$P = \frac{S_{A,True} + S_{B,True}}{S_A + S_B} \qquad (6)$$

In our study, we utilized two distinct input generation mechanisms for simulating the sorting environment: the random input generator and the seasonal pattern generator. The random input generator produces input quantities within a predefined range (e.g. 5% - 95%), ensuring variability without a discernible pattern. This method is effective in testing the robustness of the sorting system under unpredictable conditions.

Conversely, the seasonal pattern generator selects an input pattern (e.g. 10-30%, 40-60%, 70-90%) and a corresponding length (e.g. 10-12 timesteps) from a predefined range. It then maintains this pattern for the duration of its length before selecting a new pattern and length. The input phases ("Little Input," "Medium Input," and "Much

Input,") come with varying ratios of materials A and B, introducing structured periodic changes in the input flow. This approach simulates real-world scenarios where input quantities exhibit seasonal fluctuations. The left part of Fig. 4 provides an overview of all possible seasonal input patterns, while the right shows a concrete example.

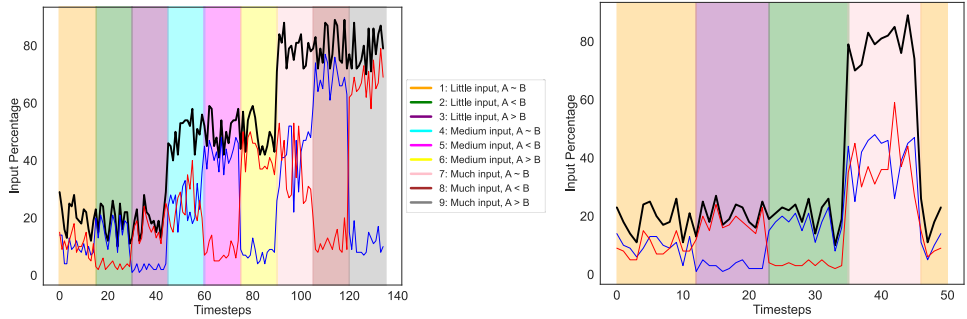

**FIGURE 4.** Overview of all nine possible seasonal input patterns (left) and a concrete example of a run with seed=42 and 50 timesteps (right). The black line presents the total input value while the colored lines illustrate the ratio of material A (blue) and B (red).

The relative weight of both factors, speed and accuracy, can be set by the operator, allowing the system to be configured for different operational goals and performance metrics. This flexibility ensures that the sorting system can be tailored to meet various production requirements and efficiency targets.

## Reinforcement Learning Formulation

To apply Reinforcement Learning, we must define an agent that learns a policy $\pi(s)$ by performing actions in our environment to maximize a cumulative reward [1]. The decision-making problem is typically formulated as a Markov Decision Process (MDP), characterized by states, actions, rewards, and transitions. In an MDP, at each time step $t$ the agent observes the current state $s_t$, selects an action $a_t$ and receives a reward $r_t$ from the environment. The state then transitions to $s_{t+1}$, and the process repeats [1].

In the proposed sorting environment, the observation space is defined as a continuous space representing the total amount of material on the input belt, normalized to the range between 0 and 1. The action space is discrete, representing different belt speeds that can be set by the agent. There are ten discrete actions corresponding to belt speeds of 10% to 100%. Each occupancy level has a defined range of belt speeds where the accuracy remains high. Beyond this range, the accuracy drops, necessitating a change in belt speed to maintain optimal sorting performance (see Fig 3, top). The reward function $R$ is designed to handle both sorting accuracy and speed operation. The reward is calculated based on the accuracy for the material currently on the belt $\alpha$, and the belt speed $v$. Both values get normalized to provide a more adequate resolution for the desired range of values. In case of the accuracy, this is all values above a threshold for minimum accuracy. All values below this will return a negative reward of -0.1 immediately. A weight for the importance of both components can be set with the reward factors $r_{acc}$ and $r_{speed}$. Optionally, a penalty term is subtracted from the results. We define reward such that there is a distinct optimal speed for each occupancy level, so the agent must dynamically adjust and find the optimal speed in each setup (see Fig. 3).

$$R = r_{\text{acc}} \cdot \left( \frac{\alpha - \text{threshold}}{1 - \text{threshold}} \right) + r_{\text{speed}} \cdot \left( \frac{v - 0.1}{0.9} \right) - \text{penalty} \tag{7}$$

The penalty can be applied to limit the total number of speed changes in input-setups that are not fully random, e.g. the seasonal input (see Fig. 4). It enhances the realism of the simulation by encouraging the agent to maintain consistent speeds within identified patterns. In real-world industrial sorting systems, frequent adjustments to conveyor belt speeds can lead to increased mechanical wear and energy consumption, along with potential disruptions to the sorting process. By simulating a scenario where speed changes are minimized, the model better reflects operational constraints and promotes efficiency. Encouraging the agent to learn and adapt to input patterns, rather than making constant speed adjustments, thus aligns the simulation more closely with practical industrial practices.

The state update process includes processing the material currently within the sorting machine, reflecting sorting accuracy influenced by belt speed and occupancy. The environment is then updated by moving the sorted material to the next station and adding new material onto the input belt. The belt speed is set based on the agent's selected action from predefined speeds. Subsequently, the sorting accuracy is recalculated, considering current belt speed and occupancy, to simulate the dynamic adjustment of sorting precision. The function then calculates the reward for the current step, providing feedback on the agent's performance. Finally, the next observation of the environment is obtained for the agent's subsequent decision-making.

## The Advanced Environment

Industrial facilities often update setups with new sensors or machine elements. The advanced sorting environment builds upon the basic environment by introducing additional upgrades that influence the sorting process, thereby increasing the complexity of the simulation. This section highlights the key modifications made in the advanced environment compared to the basic environment.

While the basic environment allowed only for discrete belt speed adjustments, we simulated a machine upgrade by introducing three sorting modes: basic sorting, positive sorting, and negative sorting. This approach, inspired by an actual technical use case [7], adds a layer of realism and strategic depth. We define correct sorting modes per ratio as:

$$\text{Basic: } \frac{1}{3} \leq \frac{A}{B} \leq 3, \quad \text{Positive: } \frac{A}{B} \geq 3, \quad \text{Negative: } \frac{A}{B} \leq \frac{1}{3} \tag{8}$$

When the correct sorting mode is selected, the machine may internally adjust frequencies or other operational parameters to enhance sorting precision and reduce noise. The action space is thus expanded to account for both belt speed and sorting mode, resulting in a total of 30 discrete actions (10 belt speeds × 3 sorting modes). The observation space in the advanced environment includes not only the total amount of input material but also the ratio of material A to material B, simulating a sensor upgrade. This additional dimension provides the agent with more detailed information about material composition, which is relevant for the choice of sorting mode. The detected ratio is encoded categorically, based on the proportion of material A to material B. The reward function remains the same, indirectly considering the chosen sorting mode as it influences the current accuracy. Sorting accuracy in the advanced environment is influenced by belt speed, occupancy, and the selected sorting mode. Additionally, noise levels vary based on the sorting mode, and incorrect mode selection incurs an indirect penalty by reducing accuracy. For example, knowing the ratio of incoming materials allows the machine to optimize its sorting mechanism, increasing accuracy by 15% and decreasing noise to a range of 0 to 5%. Conversely, selecting an incorrect sorting mode can degrade performance, reducing accuracy by 10%. Mathematically, accuracy updates including noise adjustments, are given by:

$$\alpha = \begin{cases} \min(\alpha + 0.15, 1.0) - Noise & \text{if mode correct} \\ \max(\alpha - 0.10, 0.0) - Noise & \text{if mode incorrect} \end{cases} \tag{9}$$

The range of noise is adjusted based on the choice of sorting mode:

$$Noise = \begin{cases} \text{np.random.uniform}(0.0, 0.05) & \text{if mode correct} \\ \text{np.random.uniform}(0.1, 0.15) & \text{if mode incorrect} \end{cases} \tag{10}$$

## EXPERIMENTS AND RESULTS

In this section, we conduct experiments to verify the feasibility of solving the sorting environments using reinforcement learning (RL) and to demonstrate the potential advantages of RL over traditional rule-based systems. Multiple reinforcement learning models were trained and evaluated in the sorting environments, including Proximal Policy Optimization (PPO), Deep Q-Network (DQN), and Advantage Actor-Critic models (A2C), utilizing the implementation from Stable-Baselines3 v2.2.1 [19]. Each model was trained for 100.000 timesteps. The training process involved 250 steps per episode. During testing, 50 steps per episode were taken to assess and visualize the model's performance. We set an accuracy threshold of 0.7 to serve as a quality criterion. The training environment was configured with different parameters such as the environment type (basic or advanced), input type (random or seasonal), and noise type (with or without). For scenarios using non fully random input, e.g. seasonal input, an action penalty of 0.5 was set to limit the number of speed changes, encouraging the agent to recognize input patterns and maintain consistent speeds within a pattern.

Noise was introduced by modulating the total observed amount of material on the input belt with a random perturbation. Specifically, the noise was generated as a uniform random variable within a specified range and applied to the current material amount. This method induced stronger noise effects with higher material quantity, simulating a more complex and uncertain environment as the system became more loaded. The resulting noisy observation was clamped between 0 and 1 to maintain realistic input bounds.

A rule-based agent was programmed to serve as a classical baseline for the given tasks. It initially generated a lookup table of all possible actions based on the range of possible observations and their corresponding immediate rewards. It subsequently always followed the rule to select the action that yielded the highest reward from the table. Unlike RL agents, this agent focuses solely on immediate rewards, lacking long-term decision-making and pattern recognition.

All models were trained using default parameters, with some adjustments for each model. For the PPO model, modified parameters included a discount factor of 0.85, a learning rate of 0.0007, and an entropy coefficient of 0.03. The DQN model was configured with a discount factor of 0.95. For the A2C model, we set a discount factor of 0.9, a learning rate of 0.0005, and an entropy coefficient of 0.06.

In individual experiments, the trained agents' performances were assessed and visualized in an environment with a fixed seed (seed=42), generating deterministic input. In the benchmarking experiment, each agent was trained from scratch and evaluated in a series of ten deterministic environments to benchmark its performance, recording metrics as the mean reward and standard deviations.

Having a real-world inspired environment with both options for minor changes (e.g., increasing noise) and major changes (e.g., upgrading machinery and sensors), creates optimal conditions to test how agents adapt to evolving industrial setups. The focus of these experiments was to demonstrate that all provided sorting environments with varying complexity can be effectively solved by RL agents to be used in future studies on RL adaptability.

## Results

In the following we present the results from the individual experiments and benchmarking conducted in various environmental setups. We compare the performance of different RL algorithms across four setups (A, B, C, D), detailed in Table 1 and Figures 5-7. The environments were tested under conditions without (A, B) and with noise (C, D). The input type was either random (A, C) or seasonal (B, D), with an action penalty applied in setups with seasonal input. All models were trained until they reached a performance plateau, typically after about 100,000 steps. The main observations are presented in the following:

- In all setups, agents in advanced environments achieved higher cumulative rewards than those in basic environments. In setup A (no noise, no action penalty), all algorithms achieved higher mean speeds, purity, and rewards in the advanced environment. This trend held in noisy conditions as well (see Table 1).

- In random input setups (A, C) the RL agents performed equally well as the RBA. However, in setups with seasonal input (B, D), the RL agents showed superior adaptability, especially under noisy conditions. Figures 5 and 6 highlight these differences, showing that the RL agents maintained higher purity and rewards, particularly in advanced setups. The RBA did not learn patterns, treating each input individually, resulting in lower total rewards due to frequent action penalties (Fig. 6).

- Introducing noise (C, D) generally led to a decrease in performance metrics across all models. Despite this, RL algorithms such as DQN and PPO displayed better robustness under noisy conditions, maintaining higher levels of purity and reward compared to the RBA baseline.

- The learning behavior of RL agents was significantly influenced by hyperparameters such as the learning rate, exploration rate, and discount factor. Adjustments to these parameters were crucial in optimizing agent performance, particularly in environments with higher complexity or noise.

- In setups with seasonal input, there were instances (e.g. D6, D7, Table 1) where RL agents selected different belt speeds for subsequent patterns with the same input amount as the previous pattern, highlighting the complexity of optimizing RL agents for dynamic environments.

- In some cases (Table 1), the A2C algorithm tended to select a static belt speed for the entire period, which negatively affected its performance. This collapse was more likely with certain hyperparameter settings, e.g. a lower entropy coefficient. Although a static speed might be optimal for specific occupancy levels, it proved to be suboptimal overall, leading to lower rewards compared to more dynamic strategies.

Benchmarking across different algorithms and setups confirmed the tendencies observed in individual experiments. The agents in advanced environments consistently outperformed the agents in basic environments, even under challenging conditions such as noise and action penalties. The mean total rewards illustrated that while noise and penalties generally reduced performance, in most cases the RL agents outperform the RBA baseline (Fig. 7).

**TABLE 1.** Comparison of various RL algorithms in different environmental setups (A,B,C,D). The columns represent both parameters and outcomes, including the environment type, algorithm, input type, noise level, action penalty, average speed, mean purity, mean reward, and notes. The environments are categorized as either "Basic" or "advanced," and the algorithms tested include RBA, DQN, PPO, and A2C. The input type is either random (R) or seasonal (S), and the noise levels tested are 0.0 and 0.3. The action penalty is set to either 0.0 or 0.5. The note "static" means that a static speed was chosen for the whole period. The highest values per group are in bold.

| Index | Env | Algorithm | Input | Noise | Action Penalty | Speed (Mean) | Purity (Mean) | Reward | Notes |
|-------|-----|-----------|-------|-------|----------------|--------------|---------------|--------|-------|
| A1 | Basic | RBA | R | 0.0 | 0.0 | 55 | 85 | **26.56** | |
| A2 | Basic | DQN | R | 0.0 | 0.0 | 44 | 85 | 23.9 | Fig. 2 |
| A3 | Basic | PPO | R | 0.0 | 0.0 | 53 | 85 | 26.32 | |
| A4 | Basic | A2C | R | 0.0 | 0.0 | 45 | 85 | 24.39 | |
| A5 | Adv | RBA | R | 0.0 | 0.0 | 59 | 94 | **36.48** | |
| A6 | Adv | DQN | R | 0.0 | 0.0 | 40 | 93.5 | 31.01 | |
| A7 | Adv | PPO | R | 0.0 | 0.0 | 55 | 94 | 35.29 | |
| A8 | Adv | A2C | R | 0.0 | 0.0 | 50 | 94 | 34.12 | |
| B1 | Basic | RBA | S | 0.0 | 0.5 | 72 | 83.5 | 11.95 | |
| B2 | Basic | DQN | S | 0.0 | 0.5 | 53 | 80.5 | 23.46 | |
| B3 | Basic | PPO | S | 0.0 | 0.5 | 66 | 83.5 | **28.06** | |
| B4 | Basic | A2C | S | 0.0 | 0.5 | 50 | 53 | 18.33 | static |
| B5 | Adv | RBA | S | 0.0 | 0.5 | 77 | 91.5 | 27.33 | Fig. 6 |
| B6 | Adv | DQN | S | 0.0 | 0.5 | 48 | 91.5 | 30.23 | |
| B7 | Adv | PPO | S | 0.0 | 0.5 | 68 | 91.5 | **36.85** | Fig. 5 |
| B8 | Adv | A2C | S | 0.0 | 0.5 | 44 | 91.5 | 30.61 | |
| C1 | Basic | RBA | R | 0.3 | 0.0 | 55 | 73.5 | 19.77 | |
| C2 | Basic | DQN | R | 0.3 | 0.0 | 42 | 85 | 23.34 | |
| C3 | Basic | PPO | R | 0.3 | 0.0 | 47 | 84 | **23.79** | |
| C4 | Basic | A2C | R | 0.3 | 0.0 | 48 | 83 | 23.1 | |
| C5 | Adv | RBA | R | 0.3 | 0.0 | 61 | 85 | 29.5 | |
| C6 | Adv | DQN | R | 0.3 | 0.0 | 41 | 93 | 31 | |
| C7 | Adv | PPO | R | 0.3 | 0.0 | 49 | 93.5 | **33.62** | |
| C8 | Adv | A2C | R | 0.3 | 0.0 | 50 | 92 | 32.74 | |
| D1 | Basic | RBA | S | 0.3 | 0.5 | 73 | 76 | 10.21 | |
| D2 | Basic | DQN | S | 0.3 | 0.5 | 53 | 77 | **22.98** | |
| D3 | Basic | PPO | S | 0.3 | 0.5 | 64 | 64 | 21.62 | |
| D4 | Basic | A2C | S | 0.3 | 0.5 | 50 | 53 | 18.33 | Static |
| D5 | Adv | RBA | S | 0.3 | 0.5 | 78 | 83 | 22.83 | |
| D6 | Adv | DQN | S | 0.3 | 0.5 | 49 | 89.5 | 27.96 | |
| D7 | Adv | PPO | S | 0.3 | 0.5 | 69 | 91.5 | **35.53** | |
| D8 | Adv | A2C | S | 0.3 | 0.5 | 50 | 69.5 | 25.67 | |

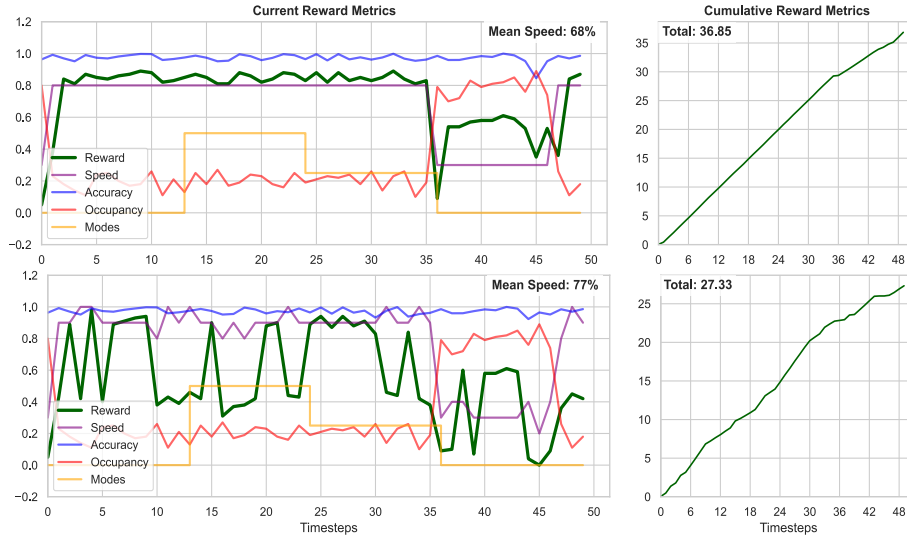

**FIGURE 5-6.** Immediate and cumulative reward metrics of the advanced sorting environment with seasonal input (for pattern, see Fig. 4, right). The top panel (Fig. 5) shows PPO Agent actions, and the bottom panel (Fig. 6) shows Rule-Based Agent actions. The left panels show the current reward metrics over 50 timesteps, including reward (green), speed (blue), accuracy (purple), occupancy (red), and sorting mode (yellow), coded as basic (0), positive sorting (0.5), and negative sorting (1.0). The right panel illustrates the cumulative reward metrics over the same timesteps.

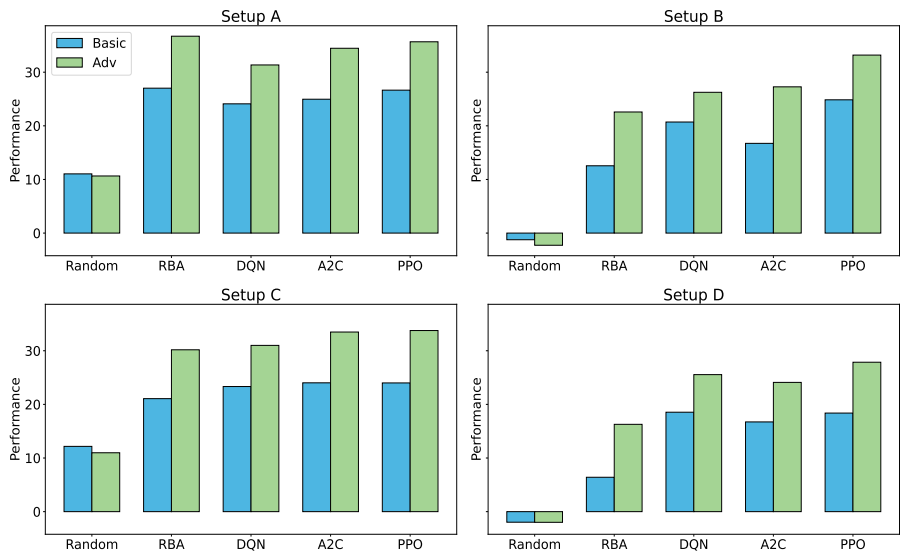

**FIGURE 7.** Comparison of benchmarking performance ("reward") of multiple RL algorithms in different setups (A, B, C, D) of the sorting environment (see Table 1). Each value depicts the mean of evaluations in ten distinct environments.

## DISCUSSION AND CONCLUSION

In this Section, we discuss the benefits and limitations of our proposed framework and outline future research directions. The primary objective of this research was to present a novel, industry-inspired environment that is publicly accessible and easily extendable for testing the adaptability of reinforcement learning (RL) agents. Our sorting environment is directly inspired by configurations commonly used in industrial settings, which involve sorting materials like metals, plastics, and paper based on various characteristics [7]. These setups aim to maximize sorting accuracy while maintaining operational efficiency. RL offers the potential to optimize these strategies, improving both accuracy and throughput.

Our environment is designed to be more extensible and applicable than most available RL benchmarks, aligning with the concept of a digital twin to enhance predictive maintenance, process optimization, and operational efficiency [20]. A significant challenge in deploying RL systems in industry is ensuring their ability to adapt to evolving real-world systems. To address this, we developed two environments: a basic version and an advanced version simulating machinery and sensor upgrades. Our findings indicate that RL agents consistently outperform rule-based agents in environments where patterns can be learned through interaction and in the presence of noise (see Fig. 7).

Our basic environment's flexibility allows for easy upgrades, providing a robust foundation for future research on agent adaptability to changing industrial setups. While the current environment includes rough estimates of plausible parameters, incorporating more precise physical parameters could enhance realism. Additionally, alternative actions beyond belt speed adjustments, such as more detailed control over sensor configurations, may be more relevant in certain setups. Future research should investigate transfer learning, meta-learning, and continual learning, where agents trained in simpler environments are tested in more complex settings, thereby reducing the need for retraining [21]. Examining adjustments in operational goals, such as minimum quality thresholds, could also further assess the real-time adaptability of RL agents.

In conclusion, our study demonstrates the potential of RL to optimize industrial processes, offering a framework for deploying RL agents in real-world settings and enhancing both efficiency and quality in production lines. While promising, further research is needed to address the challenges of adapting RL to dynamic industrial environments, ensuring that these systems can meet the evolving demands of modern industry.

## ACKNOWLEDGEMENT

This research received external funding from the German Federal Ministry for Economic Affairs and Climate Action through the grant "EnSort".

## REFERENCES

1. Sutton, R.S., Barto, A.G.: Reinforcement learning: an introduction. The MIT Press, Cambridge, Massachusetts (2018).
2. Mnih, V., Kavukcuoglu, K., Silver, D., Graves, A., Antonoglou, I., Wierstra, D., Riedmiller, M.: Playing Atari with Deep Reinforcement Learning, (2013).
3. Dulac-Arnold, G., Levine, N., Mankowitz, D.J., Li, J., Paduraru, C., Gowal, S., Hester, T.: Challenges of real-world reinforcement learning: definitions, benchmarks and analysis. Mach. Learn. 110, 2419–2468 (2021).
4. Pendyala, A., Dettmer, J., Glasmachers, T., Atamna, A.: ContainerGym: A Real-World Reinforcement Learning Benchmark for Resource Allocation. In: Machine Learning, Optimization, and Data Science. pp. 78–92. Springer Nature Switzerland, Cham (2024).
5. Dogru, O., Xie, J., Prakash, O., Chiplunkar, R., Soesanto, J., Chen, H., Velswamy, K., Ibrahim, F., Huang, B.: Reinforcement Learning in Process Industries: Review and Perspective. IEEECAA J. Autom. Sin. 11, 283–300 (2024).
6. Arents, J., Greitans, M.: Smart Industrial Robot Control Trends, Challenges and Opportunities within Manufacturing. Appl. Sci. 12, 937 (2022).
7. Kroell, N., Maghmoumi, A., Dietl, T., Chen, X., Küppers, B., Scherling, T., Feil, A., Greiff, K.: Towards digital twins of waste sorting plants: Developing data-driven process models of industrial-scale sensor-based sorting units by combining machine learning with near-infrared-based process monitoring. Resour. Conserv. Recycl. 200, 107257 (2024).
8. Schulman, J., Wolski, F., Dhariwal, P., Radford, A., Klimov, O.: Proximal Policy Optimization Algorithms, http://arxiv.org/abs/1707.06347, (2017).
9. Mnih, V., Badia, A.P., Mirza, M., Graves, A., Lillicrap, T.P., Harley, T., Silver, D., Kavukcuoglu, K.: Asynchronous Methods for Deep Reinforcement Learning, http://arxiv.org/abs/1602.01783, (2016).
10. Jerfel, G., Griffiths, T.L., Grant, E., Heller, K.: Reconciling meta-learning and continual learning with online mixtures of tasks. Advances in neural information processing systems, (2019).
11. Maus, T., Zengeler, N., Sänger, D., Glasmachers, T.: Volume Determination Challenges in Waste Sorting Facilities: Observations and Strategies. Sensors. 24, 2114 (2024).
12. Annaswamy, A.M.: Adaptive Control and Intersections with Reinforcement Learning. Annu. Rev. Control Robot. Auton. Syst. 6, 65–93 (2023).

13. Mattera, G., Caggiano, A., Nele, L.: Optimal data-driven control of manufacturing processes using reinforcement learning: an application to wire arc additive manufacturing. J. Intell. Manuf. (2024).
14. Stranieri, F., Stella, F.: A Deep Reinforcement Learning Approach to Supply Chain Inventory Management. Presented at the European Workshop on Reinforcement Learning, Milan, Italy August 13 (2022).
15. Zhang, L., Yan, Y., Hu, Y., Ren, W.: Reinforcement learning and digital twin-based real-time scheduling method in intelligent manufacturing systems. IFAC-Pap. 55, 359–364 (2022).
16. Khdoudi, A., Masrour, T., El Hassani, I., El Mazgualdi, C.: A Deep-Reinforcement-Learning-Based Digital Twin for Manufacturing Process Optimization. Systems. 12, 38 (2024).
17. Hein, D., Depeweg, S., Tokic, M., Udluft, S., Hentschel, A., Runkler, T.A., Sterzing, V.: A Benchmark Environment Motivated by Industrial Control Problems. In: 2017 IEEE Symposium Series on Computational Intelligence (SSCI). pp. 1–8 (2017).
18. Towers, M., Terry, J.K., Kwiatkowski, A., Balis, J.U., Cola, G., Deleu, T., Goulão, M., Kallinteris, A., KG, A., Krimmel, M., Perez-Vicente, R., Pierré, A., Schulhoff, S., Tai, J.J., Tan, A.J.S., Younis, O.G.: Gymnasium, https://zenodo.org/records/11232524, (2024).
19. Raffin, A., Hill, A., Gleave, A., Kanervisto, A., Ernestus, M., Dormann, N.: Stable-Baselines3: Reliable Reinforcement Learning Implementations. J. Mach. Learn. Res. 22, 1–8 (2021).
20. Siraskar, R., Kumar, S., Patil, S., Bongale, A., Kotecha, K.: Reinforcement learning for predictive maintenance: a systematic technical review. Artif. Intell. Rev. 56, 12885–12947 (2023).
21. Zhu, Z., Lin, K., Jain, A.K., Zhou, J.: Transfer Learning in Deep Reinforcement Learning: A Survey. IEEE Trans. Pattern Anal. Mach. Intell. 45, 13344–13362 (2023).